\documentclass[lettersize,journal]{IEEEtran}
\usepackage{amsmath,amsfonts}
\usepackage{algorithm, algorithmic}
\usepackage{array}
\usepackage[caption=false,font=normalsize,labelfont=sf,textfont=sf]{subfig}
\usepackage{textcomp}
\usepackage{stfloats}
\usepackage{url}
\usepackage{verbatim}
\usepackage{graphicx}
\usepackage{cite}
\usepackage{float}
\usepackage{stfloats} 
\usepackage{amssymb}
\usepackage{amsthm}
\usepackage{mathrsfs}
\usepackage{pifont}
\usepackage{multirow}
\usepackage{multicol}
\usepackage{tabulary}
\usepackage{booktabs}
\usepackage{color,xcolor}
\usepackage{CJK}
\usepackage{threeparttable}
\usepackage{makecell}
\usepackage{utfsym}
\usepackage[
pdfauthor={derajan},
pdftitle={How to do this},
pdfstartview=XYZ,
bookmarks=true,
colorlinks=true,
linkcolor=blue,
urlcolor=blue,
citecolor=blue,
pdftex,
bookmarks=true,
linktocpage=true,   
hyperindex=true
]{hyperref}
\definecolor{gold}{RGB}{255,215,0}
\definecolor{silver}{RGB}{192,192,192}
\definecolor{bronze}{RGB}{205,127,50}
\begin{document}
\title{PLE-SLAM: A Visual-Inertial SLAM Based on Point-Line Features and Efficient IMU Initialization}

\author{{Jiaming He, Mingrui Li, Yangyang Wang, \IEEEmembership{Member, IEEE}, Hongyu Wang, \IEEEmembership{Member, IEEE}
\vspace{-0.8cm}
\thanks{ This work was supported in part by the National Natural Science Foundation of China under Grants 61671103 and is part by the Science and Technology Innovation Funds of Dalian under Grants 2022JJ11CG002. \textit{(Corresponding author: Hongyu Wang.)}}

\thanks{Jiaming He, Mingrui Li and Hongyu Wang are with the School of Information and Communication Engineering, Dalian University of Technology, Dalian 116024, China  (e-mail: hjm1187961271@mail.dlut.edu.cn; 2905450254@mail.dlut.edu.cn; whyu@dlut.edu.cn).

Yangyang Wang is with the School of Information Science and Technology, Dalian
Maritime University, Dalian 116026, China (e-mail: yyw@dlmu.edu.cn)}}}

\markboth{IEEE TRANSACTIONS ON INSTRUMENTATION AND MEASUREMENT,~Vol.~14, No.~8, August~2021}%
{He \MakeLowercase{\textit{et al.}}: PLE-SLAM: A Visual-Inertial SLAM Based on Point-Line Features and Efficient IMU Initialization}


\maketitle

\begin{abstract}
Visual-inertial SLAM is crucial in various fields, such as aerial vehicles, industrial robots, and autonomous driving. The fusion of camera and inertial measurement unit (IMU) makes up for the shortcomings of a signal sensor, which significantly improves the accuracy and robustness of localization in challenging environments. This article presents PLE-SLAM, an accurate and real-time visual-inertial SLAM algorithm based on point-line features and efficient IMU initialization. First, we use parallel computing methods to extract features and compute descriptors to ensure real-time performance. Adjacent short line segments are merged into long line segments, and isolated short line segments are directly deleted. Second, a rotation-translation-decoupled initialization method is extended to use both points and lines. Gyroscope bias is optimized by tightly coupling IMU measurements and image observations. Accelerometer bias and gravity direction are solved by an analytical method for efficiency. To improve the system's intelligence in handling complex environments, a scheme of leveraging semantic information and geometric constraints to eliminate dynamic features and A solution for loop detection and closed-loop frame pose estimation using CNN and GNN are integrated into the system. All networks are accelerated to ensure real-time performance. The experiment results on public datasets illustrate that PLE-SLAM is one of the state-of-the-art visual-inertial SLAM systems.
\end{abstract}

\vspace{-0.1cm}
\section{Introduction}
\IEEEPARstart{S}{imultaneous} localization and mapping (SLAM) is the basis for realizing autonomous navigation of mobile robots and has received close attention from researchers in recent years. Although vision-based SLAM systems\cite{SVO,Forster17troSVO,ORB-SLAM2,DSO} are popular for their low hardware costs and good localization performance, they still have problems with challenging scenes such as fast motion, dynamic illumination, viewpoint change, and occlusion, etc. Inertial navigation is fearless of visual challenges, but cumulative drift makes it difficult to achieve long-term SLAM. Visual-inertial SLAM effectively makes up for the shortcomings of a single sensor by combining high-frequency inertial measurements and visual geometric information from different sensors. Nowadays, VI-SLAM and VIO systems\cite{OKVIS,VINS-Mono,VI-ORB-SLAM,vidso,SVOPro,ORB-SLAM3} are widely used in lots of practical products and have broad application prospects. This paper proposes a complete visual-inertial SLAM system consisting of a stereo camera and a signal IMU.
\par
Most existing visual-inertial SLAM systems are based on only point features and track through optical flow or descriptor matching. Due to the low calculation cost and high tracking efficiency of manual feature points\cite{orb}, point-based VI-SLAM systems are extensively researched and developed. However, these methods may suffer from issues in low-texture environments, such as tracking drift and failure of visual pose estimation. Although IMU can compensate for the pose estimation error caused by the lack of point features to a certain extent, long-term feature loss will cause the IMU accumulated error to be too unable to be effectively eliminated. Some methods\cite{plsvo,plslammono,plslam,airvo,plvio,pl-vins,tvtplslam,eplfvins} introduce line features that are common in artificial scenes in vision-based methods to provide more geometric constraints. Line features are more prevalent in artificial scenes and more robust to lighting and viewpoint change. We then incorporate line features\cite{edlines} into a stereo-inertial SLAM system. The proposed method uses two line segment projection constraints and one point projection constraints for visual bundle adjustment, dramatically improving tracking performance in challenging environments.
\par
In addition to stable tracking, the IMU initialization is also crucial for a visual-inertial system. Accurate initial values enable faster convergence of optimization functions. The IMU parameters that need to be estimated by VI-SLAM or VIO mainly include gyroscope bias, accelerometer bias, gravity vector, velocity and scale. Due to the constraints of the stereo extrinsic matrix in this work, we fix the scale to 1.0. IMU initialization is mainly divided into tightly-coupled methods\cite{tightclosed,openvins} and loosely-coupled methods\cite{vinsintiliaztion,inertialonly,analytical}. Tightly-coupled methods estimate rotation with raw gyroscope measurements and calibrated camera IMU extrinsic matrix and then get the initialization parameters by solving a large matrix containing camera and accelerometer observations. These methods are time-consuming and ignore bias, which may lead to bad initial values on consumer-grade IMU devices. Some loosely-coupled methods estimate rotation using camera and gyroscope measurements, respectively, and then estimate inertial parameters iteratively. However, purely visual pose estimation is not robust enough for rotation and insufficient features. Inspired by \cite{drt} and \cite{efficient}, we convert point observations under a multi-camera system into line observations under a signal camera and propose an efficient visual-inertial initialization algorithm. The method estimates gyroscope bias utilizing 2D point-line observations and gyroscope measurements and solves other initial state variables through an analytical solution.
\par
In this work, we further employ deep neural networks (DNN) to improve the intelligence level of the system. First, a dynamic feature elimination module is integrated into the system to improve the localization accuracy in dynamic scenes. Second, the loop closing thread is modified to use deep features and matching technology to adapt to challenging environments and obtain more accurate loop frame pose estimation. All DNN-based modules are accelerated with TensorRT to ensure real-time. In summary, the main contributions of this study are as follows:
\par
\begin{itemize}
    \item [1)]
     To improve the accuracy and robustness of SLAM systems in challenging environments, we propose a visual-inertial SLAM based on point-line features. We merge or suppress unstable lines to ensure the stability of line features during tracking. The point and line features are tightly coupled in tracking and bundle adjustment processes.
\end{itemize}
\par
\begin{itemize}
    \item [2)]
    An efficient visual-inertial initialization method that contains a hybrid process of iterative and analytical solutions is presented. The gyroscope bias is directly optimized by a point-line rotation-only solution with visual observations. Accelerometer bias and gravity are solved by an analytical solution, which is faster than an iterative solution. All inertial parameters are refined through MAP in the back-end.
\end{itemize}
\par
\begin{itemize}
    \item [3)]
    To better adapt to complex scenes, we employ DNN-based methods in our system, which contains a dynamic feature elimination module and a modified loop closing thread using SuperPoint\cite{superpoint} and SuperGlue\cite{superglue}. The employment of DNN-based methods improves the stability and localization accuracy of the system in long-term operation in complex environments.
\end{itemize}
\par
All these innovations and engineering implementations of code make PLE-SLAM one of the state-of-the-art visual-inertial SLAM methods in complex environments. We release the code at \url{https://github.com/HJMGARMIN/PLE-SLAM} for the benefit of the community.
\section{Related Work}
\subsection{Point-Line Based SLAM}
Inspired by many open-source projects, visual SLAM has achieved tremendous development. Many researchers have introduced line segment features based on the traditional point-based SLAM framework\cite{ORB-SLAM2,SVO} to improve system accuracy and robustness. \cite{plsvo} proposed a semi-direct monocular by combining points and line segments. \cite{plslammono} added point and line features to monocular SLAM to obtain more accurate results in indoor environments, and \cite{plslam} further extended it to the stereo system. 
\par
With the assistance of IMU, visual-inertial SLAM shows better performance than pure visual SLAM. Nowadays, there are many excellent visual-inertial SLAM systems based on point-line fusion. Xu et al.\cite{airvo} innovatively combined DNN-based points and traditional line segments and merged adjacent short line segments to improve the robustness in dynamic illumination scenes. PL-VIO\cite{plvio} proposed a point-line VIO based on \cite{VINS-Mono}. Fu et al.\cite{pl-vins} improved the efficiency of PL-VIO by modifying LSD\cite{lsd} algorithm. \cite{tvtplslam} uses IMU information to guide optical flow tracking and line feature matching. \cite{eplfvins} proposed line optical feature tracking, which replaces the descriptor-based matching method and dramatically improves the speed of system.
\vspace{-0.2cm}
\subsection{Visual-Inertial Initialization}
Because initial state variables play a crucial role in the iterative optimization processes, a robust and accurate initialization method is necessary to VINS. \cite{vinsintiliaztion,VI-ORB-SLAM,analytical,fast} all solve gyroscope bias iteratively at the beginning of initialization. Among them, \cite{vinsintiliaztion} considers that the accelerometer bias is tiny relative to gravity and can be ignored, while \cite{VI-ORB-SLAM,analytical,fast} estimate accelerometer bias, scale and gravity direction via Singular Value Decomposition (SVD) or Lagrange multiplier method. Recently, \cite{inertialonly} utilizes maximum a posteriori (MAP) to estimate all inertial state variables together, considering the uncertainty of IMU measurements. The above methods all rely on the sparse point cloud obtained by SfM, which is not accurate enough when facing rapid rotation. \cite{openvins} estimates rotation using calibrated extrinsic matrix and gyroscope measurements and then provides a closed-form solution for estimating velocity and gravity direction. However, the method does not consider the gyroscope bias, resulting in inaccurate rotation estimation on consumer-grade IMU. \cite{drt} proposed a rotation-translation-decoupled solution for visual-inertial initialization, which can obtain gyroscope bias from raw image observations and rotation pre-integration without SfM.
\vspace{-0.2cm}
\subsection{SLAM Integrating Deep Neural Network}
Deep learning technology has rapidly progressed in the past few years, and deployment and computing costs have been significantly reduced. Some SLAM algorithms use deep neural networks to replace one or more modules in the original framework, which improves system accuracy and robustness in corner cases while ensuring real-time. \cite{okvis2} employed a lightweight semantic segmentation network to remove key points with large depth on the sky or clouds. \cite{dynamicvins,dgmvins,ovd,cfp} integrate semantic segmentation or object detection networks to obtain semantic information in the scene, and then combine geometric constraints to remove dynamic features. \cite{airvo} and \cite{gcnv2} use DNN-based key points to replace manual key points and achieve certain accuracy improvements. \cite{srvio} utilizes pre-trained CNN networks to compensate IMU noise to solve the problem of integration divergence problem. The integration of deep neural networks gives SLAM systems a certain degree of intelligence to deal with complex situation, especially when facing moving objects or dynamic illumination.

\section{Methodology}
\subsection{System Overview}
\begin{figure*}[htbp]
\centering
\includegraphics[width=\textwidth]{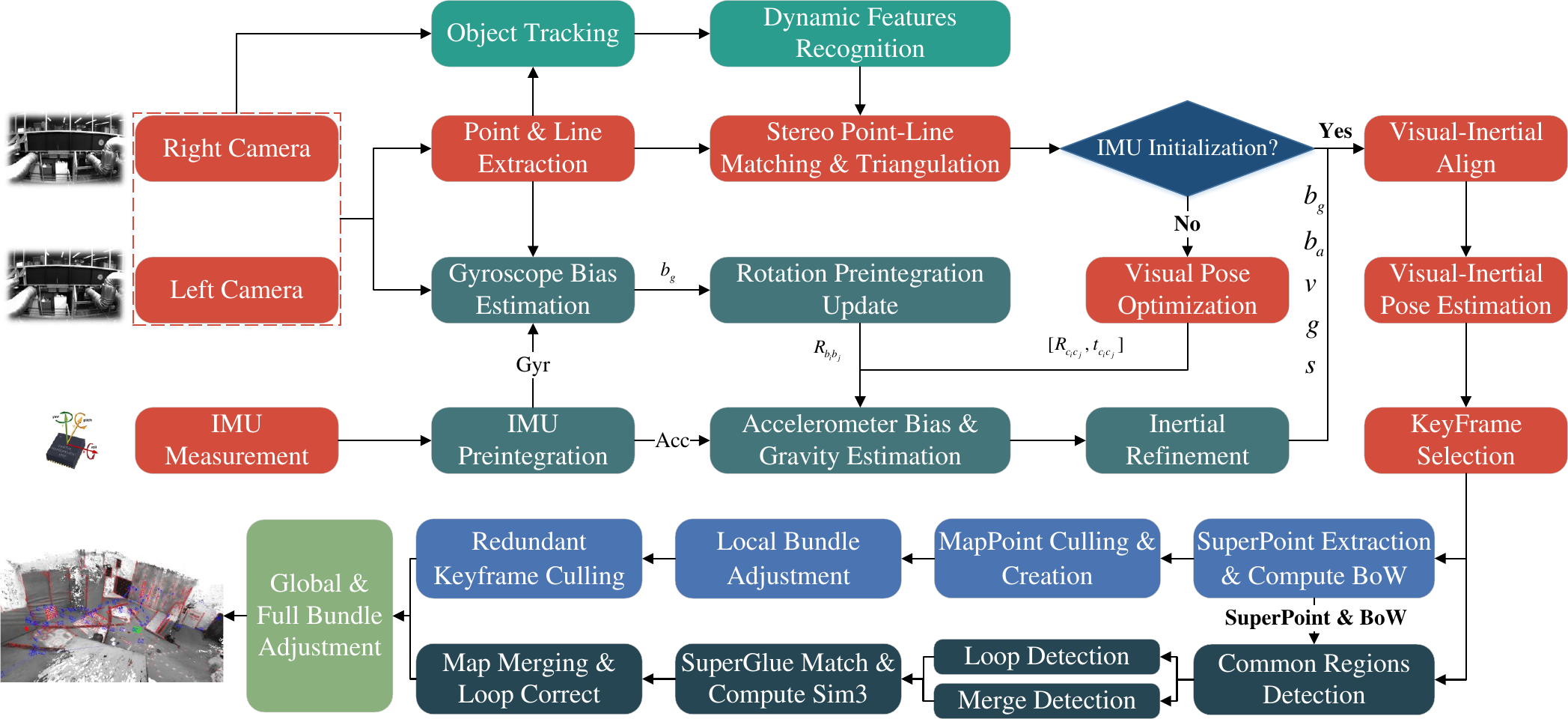}
\caption{The framework of the proposed system. We extract point and line segment features from stereo images. The 2D feature observations and rotation pre-integration are combined to estimate gyroscope bias by a rotation-only solution. Accelerometer bias and gravity direction are solved by an analytical solution, which makes the initialization process faster than iterative methods. In back-end, DNN-based features and matching methods are used to detect loop candidate frames and estimate the relative pose of loop frames. To improve the robustness to dynamic environments, a dynamic feature elimination thread is performed in parallel with tracking thread by combining semantic information and geometric constraints. The dense point cloud map in the figure is constructed using the left camera keyframe image and the depth map obtained by semi-global block matching.}
\label{framework}
\end{figure*}

As shown in Fig. \ref{framework}, the proposed PLE-SLAM takes stereo images and single IMU measurements as input. The system is based on ORB-SLAM3 and has a total of four threads, namely tracking, local mapping, loop closing and dynamic feature elimination. 
\par
Point and line features are extracted in parallel and tightly coupled with gyroscope information to estimate gyroscope bias directly, which will be used to update rotation pre-integration. The accelerometer bias and gravity direction are solved by an analytical solution conducted by the pre-integration of velocity and position. After completing the initial value estimation, all inertial variables are refined in a MAP framework. In the dynamic feature elimination thread, we employ an object tracking method to compensate for the missed detection of object detection network. The semantic information and geometric constraints are combined to recognize dynamic features. In the back-end, we utilize SuperPoint\cite{superpoint} and bag-of-words for loop closure, and use SuperGlue\cite{superglue} to match key points between loop frames. The method improves the recall and accuracy of pose estimation between loop frames in complex environments. All static point and line segment geometric constraints are used in visual bundle adjustment. Detailed technical introduction will be introduced in the subsequent parts of this section.
\vspace{-0.1cm}
\subsection{Feature Processing}
\subsubsection{Feature Extraction and Matching}
In terms of point features, we use a feature point extraction and matching strategy similar to \cite{ORB-SLAM2}. The difference is that we process all pyramid images and meshes in parallel via Thread Building Blocks (TBB), significantly improving the speed of feature extraction and descriptor computation by about nine times. The method is insensitive to the number of feature points. When the number of feature points doubles, the time consumption increases slightly.
\par
To keep up with the speed of feature point extraction as much as possible, line extraction is based on EDLines\cite{edlines} for efficiency. Compared to LSD\cite{lsd}, EDLines is nearly ten times faster with similar extraction quantities and qualities. In terms of matching, we select candidate matched lines according to their distances from origin and their inclinations, and then filter the most suitable candidate lines based on the descriptor.
\subsubsection{Unstable Lines Fusion and Suppression}
In visual-inertial SLAM, longer line segments mean more stable tracking effect. Because the line segments extracted by EDLines may be divided into multiple small segments due to factors such as image noise, we fuse two segments $L_1$ and $L_2$ based on the following conditions:
\par
\begin{itemize}
    \item 
     The angle difference between two line segments is less than 1 degree.
\end{itemize}
\par
\begin{itemize}
    \item 
     The projections of two segments are not overlapping, but the distance between the closest endpoints of them is less than 10 pixels.
\end{itemize}
\par
\begin{itemize}
    \item 
     The vertical distances from each group of startpoint, midpoint and endpoint of a line segment to another line segment are all less than 3 pixels.
\end{itemize}
\par
\begin{figure}[htbp]
\centering
\setlength{\abovecaptionskip}{-0.2cm}
\includegraphics[width=\columnwidth]{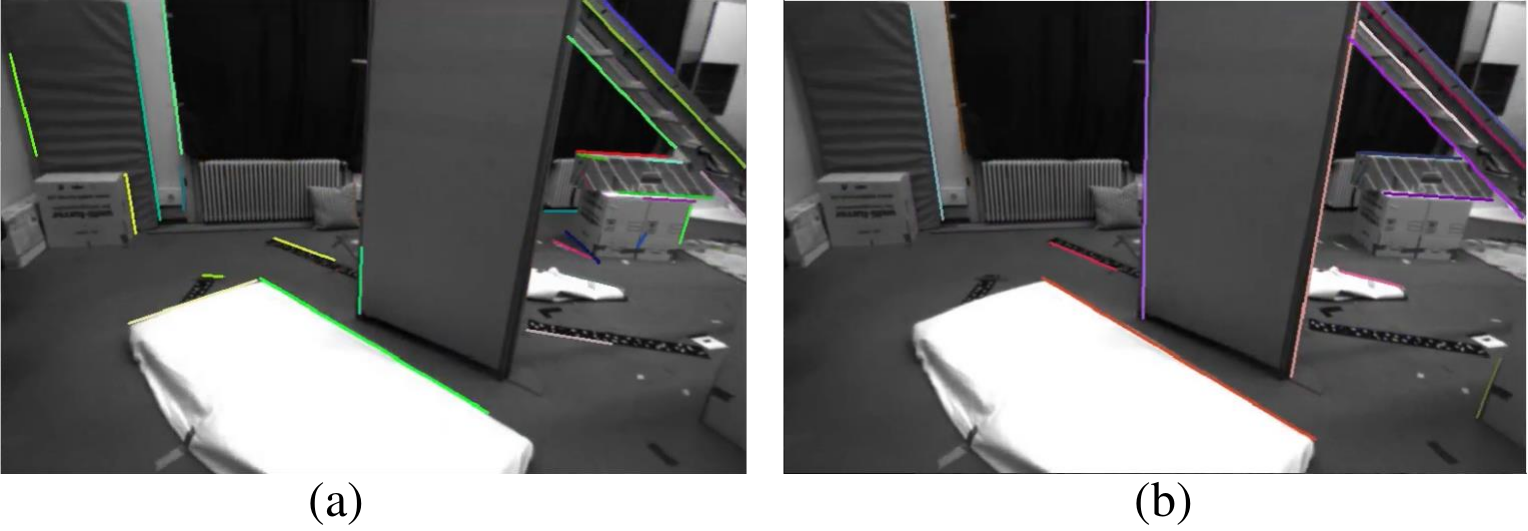}
\caption{Line segments extracted by EDLines (a) and by our proposed method (b). In order to make it easier to observe the effect, different line segments are drawn in different colors. Obviously, our method effectively merges and filters short line segments and obtains longer line segments that are more stable to track.}
\label{merge}
\end{figure}

Line segments are merged only when all three conditions are met. We set the length threshold of the shortest line segment according to the image size and line segments shorter than the threshold will be filtered out. The comparison results before and after merging and filtering are shown in Fig. \ref{merge}.
\vspace{-0.2cm}
\subsection{IMU Initialization}
\subsubsection{IMU Measurement Model and Bias Update}
IMUs can typically output acceleration $\widetilde{a}(t)$ and angular velocity $\widetilde{\omega}(t)$ measurements in an inertial reference frame. Due to the influence of sensor errors, the relationship between sensor measurement values and real values is as follows:
\begin{equation}
    \begin{aligned}
        \omega(t)&=\widetilde{\omega}(t)-b_g-\eta_{gd}\\
        a(t)&=\widetilde{a}(t)-b_a-\eta_{ad}
    \end{aligned}
\end{equation}
where $b_g$ and $b_a$ represent gyroscope bias and accelerometer bias respectively. $\eta_{gd}$ and $\eta_{ad}$ denote random noise of IMU and are assumed to follow a zero-mean Gaussian distribution.
\par
Assume that the timestamps corresponding to two image frames $c_i$ and $c_j$ are $t_i$ and $t_j$, the IMU pre-integration can be expressed by the method proposed in \cite{manifold}.
\begin{equation}
    \begin{aligned}
        \Delta \textbf{R}_{b_ib_j}&=\prod_{k=i}^{j-1}Exp((\tilde{\omega}_k-b_g-\eta_{gd})\Delta t_{ij})\\
        \Delta \textbf{v}_{b_ib_j}&=\textbf{R}_{b_i}^{T}(\textbf{v}_{b_j}-\textbf{v}_{b_i}-\textbf{g} \Delta t_{ij})\\
        \Delta \textbf{p}_{b_ib_j}&=\textbf{R}_{b_i}^{T}(\textbf{p}_{b_j}-\textbf{p}_{b_i}-\textbf{v}_{b_i} \Delta t_{ij}-\frac{1}{2}\textbf{g}{\Delta t_{ij}}^2)
    \end{aligned}
    \label{pre-integration}
\end{equation}
where $\textbf{R}_{b}$, $\textbf{v}_{b}$ and $\textbf{p}_{b}$ are rotation, velocity and translation at the corresponding moment, respectively. $\textbf{g}$ is the gravity in the body frame. $\Delta \textbf{R}_{b_ib_j}$, $\Delta \textbf{v}_{b_ib_j}$ and $\Delta \textbf{p}_{b_ib_j}$ denote the pre-integration of rotation, velocity and translation respectively. $\Delta t_{ij}$ is the difference between $t_j$ and $t_i$.
\par
Although the bias can be considered a constant during the pre-integration, it is a state quantity for optimization in visual-inertial systems. After the bias is updated, recalculating the pre-integration is time-consuming. Therefore, we follow the approach in \cite{manifold} and give a formula that directly calculates the pre-integration when the bias changes.
\begin{equation}
    \begin{aligned}
        \Delta \overline{{\textbf{R}}}_{b_ib_j}&\triangleq \Delta{\textbf{R}}_{b_ib_j}\cdot Exp(J^{b_g}_{\Delta \textbf{R}_{ij}}\delta b_g)\\
        \Delta \overline{{\textbf{v}}}_{b_ib_j}&\triangleq \Delta{\textbf{v}}_{b_ib_j}+J^{b_g}_{\Delta \textbf{v}_{ij}}\delta b_g+J^{b_a}_{\Delta \textbf{v}_{ij}}\delta b_a\\
        \Delta \overline{{\textbf{p}}}_{b_ib_j}&\triangleq \Delta{\textbf{p}}_{b_ib_j}+J^{b_g}_{\Delta \textbf{p}_{ij}}\delta b_g+J^{b_a}_{\Delta \textbf{p}_{ij}}\delta b_a
    \end{aligned}
    \label{update}
\end{equation}
where $\delta b = [\delta b_g\ \ \delta b_a]^T$ is the change of IMU bias. Taking the gyroscope bias as an example, the new bias $\overline{b}_g$ can be expressed as $\overline{b}_g=b_g+\delta b_g$. $J^{b_g}$ and $J^{b_a}$ represent the Jacobian matrix of the corresponding pre-integration relative to $b_g$ and $b_a$. $ \Delta \overline{{\textbf{R}}}_{b_ib_j}$, $\Delta \overline{{\textbf{v}}}_{b_ib_j}$ and $\Delta \overline{{\textbf{p}}}_{b_ib_j}$ are the updated pre-integration.
\subsubsection{Gyroscope Bias Estimation}
If there is a accurate initial value, the rotation between images can be directly optimized through 2D feature observations. \cite{he2023rotation} uses the rotation estimated by gyroscope measurements as the initial value and extend the method in \cite{direct} to visual-inertial initialization. We review the specific principles in this paragraph. For a pair of matching points $p_i$ and $p_j$ in two frames, their unit vectors from camera optical center are defined as $f_i$ and $f_j$. The normal vector corresponding to the epipolar plane can be expressed as $n_k=[f_i^k]_\times \textbf{R}_{c_i c_j}f^k_j$, where $k$ denotes the $k^{th}$ pair. Because multiple normal vectors must remain coplanar, the martrix $\textbf{N}=[n^1,\cdots,n^n]$ must be dissatisfied with rank, that is, the minimum eigenvalue $\lambda_{\textbf{M}_{ij},min}$ of $\textbf{M}=\textbf{N}\textbf{N}^T$ is zero. The rotation transformation between the body frame and the camera frame is $\textbf{R}_{c_i c_j}=\textbf{R}_{bc}^T\textbf{R}_{b_i b_j}\textbf{R}_{bc}$. According to the calibrated extrinsic rotation matrix $\textbf{R}_{bc}$ and rotation pre-integration in Eq. (\ref{update}), the detailed expression of $\textbf{M}$ can be written as:
\begin{equation}
\begin{aligned}
    \textbf{M}_{ij}=\sum^{n}_{k=1}([f_i^k]_\times \textbf{R}_{bc}^T\Delta \textbf{R}_{ij}Exp(J^{b_g}_{\Delta \textbf{R}_{ij}}\delta b_g)\textbf{R}_{bc}f^k_j)\\([f_i^k]_\times \textbf{R}_{bc}^T\Delta \textbf{R}_{ij}Exp(J^{b_g}_{\Delta \textbf{R}_{ij}}\delta b_g)\textbf{R}_{bc}f^k_j)^T
\end{aligned}
\end{equation}
For the point pair $(i,j)\in \mathbb{P}$, the gyroscope bias is obtained by iteratively optimizing the function shown in Eq. (\ref{cost1}). Initial IMU bias is assumed to be zero vector, so $\delta b_g^*$ is actually the result of gyroscope bias.
\begin{equation}
    \begin{aligned}
        \delta b_g^*&=\mathop{\arg\min}_{\delta b_g}\lambda\\
        with\ \  \lambda&=\sum_{(i,j)\in \mathbb{P}}\lambda_{\textbf{M}_{ij},min}
    \end{aligned}
    \label{cost1}
\end{equation}
\par
Line segments in an image can be expressed through the normal vector of the polar plane where the line segment lies in the form of  Pl{\"u}cker line-vectors. The first three dimensions are the direction vector of the normal vector passing through the midpoint of the line segment, and the last three dimensions are the moment of the normal vector, which is expressed as the cross product of a vector from origin to the midpoint of the line and the normal vector's direction. 
\vspace{0.1cm}
\begin{equation}
    L_i=\begin{pmatrix}
    l_i \\
    d_i \times l_i
\end{pmatrix}
\label{line}
\end{equation}
where $d_i$ represents the vector from origin to midpoint of the line. According to the epipolar constraints of the line in \cite{efficient}, we can construct a generalized epipolar plane normal vector $g_k$ similar to $n_k$ mentioned earlier.
\begin{equation}
    g_k=\begin{pmatrix}
     [l_i^k]_\times \textbf{R}_{c_i c_j}l_j^k  \\
     l_i^k([d_i^k]_\times \textbf{R}_{c_i c_j}-\textbf{R}_{c_i c_j}[d_j^k]_\times)l_j^k
\end{pmatrix}
\end{equation}
\par
From previous works we know that the rank of $\textbf{G}=[g^1,\cdots,g^m]$ is at most 3 and we can construct $\textbf{H}=\textbf{G}\textbf{G}^T$ and solve for the rotation by minimizing its minimum eigenvalue. So far, the point-based solution is achieved by minimizing the smallest eigenvalue of $\textbf{M}$, and the line-based solution is achieved by minimizing the smallest eigenvalue of $\textbf{H}$. Since both $\textbf{M}$ and $\textbf{H}$ are real symmetric matrices, their eigenvalues are non-negative. We can optimize the function as shown in Eq. (\ref{pointline}), which combines point and line feature in a rotation-only solution to directly gyroscope bias and is solved by calculation tools in \cite{opengv}. Let $\lambda_{\textbf{H}_{i^{'}j^{'}},min}$ be the smallest eigenvalue of $\textbf{H}$, which can be expressed in closed-form by applying Ferrari's solution. The square root in Eq. (\ref{pointline}) is used to make the iterative optimization converge faster. The final result is used to update the rotation pre-integration in Eq. (\ref{update}).
\begin{figure*}[htbp]
    \begin{equation}
     \begin{aligned}
    \textbf{H}_{ij}=\sum^{m}_{k=1}   
    \begin{pmatrix}
        [l_i^k]_\times \textbf{R}_{bc}^T\textbf{R}_{c_i c_j}\textbf{R}_{bc}l_j^k \\
            (l_i^k)^T([\textbf{R}_{bc}^T \Delta \textbf{R}_{ij}d^k_i]_\times Exp(J^{b_g}_{\Delta \textbf{R}_{ij}}b_g)\textbf{R}_{bc}-Exp(J^{b_g}_{\Delta \textbf{R}_{ij}}b_g)\textbf{R}_{bc}[d_j^k]_\times)l_j^k
    \end{pmatrix} \\
    {\begin{pmatrix}
        [l_i^k]_\times \textbf{R}_{bc}^T\textbf{R}_{c_i c_j}\textbf{R}_{bc}l_j^k \\
            (l_i^k)^T([\textbf{R}_{bc}^T \Delta \textbf{R}_{ij}d_i^k]_\times Exp(J^{b_g}_{\Delta \textbf{R}_{ij}}b_g)\textbf{R}_{bc}-Exp(J^{b_g}_{\Delta \textbf{R}_{ij}}b_g)\textbf{R}_{bc}[d_j^k]_\times)l_j^k
    \end{pmatrix}}^T
    \end{aligned}
\end{equation}
\vspace{-0.3cm}
\end{figure*}
\begin{equation}
    \begin{aligned}
        \delta b_g^*&=\mathop{\arg\min}_{\delta b_g}\lambda\\
        with\ \  \lambda=\sum_{\substack{(i,j)\in \mathbb{P}\\(i^{'},j^{'})\in \mathbb{L}}}&\sqrt{\lambda_{\textbf{M}_{ij},min}}+\sqrt{\lambda_{\textbf{H}_{i^{'}j^{'}},min}}
    \end{aligned}
    \label{pointline}
\end{equation}
\subsubsection{Analytical Solution for  Accelerometer Bias and Gravity Direction}
We are using s stereo camera, so the scale of translation can be obtained from the baseline. In this section, we compute the accelerometer bias and gravity direction analytically with a similar method to \cite{analytical} and \cite{fast}. Given the extrinsic rotation $\textbf{R}_{cb}$ and translation $\textbf{t}_{cb}$ between camera and IMU, we can translate the poses in camera frame to body frame as:
\begin{equation}
    \begin{aligned}
        \textbf{R}_{b_i}&=\textbf{R}_{c_i}\textbf{R}_{cb}\\
        \textbf{p}_{b_i}&=s\textbf{p}_{c_i}+\textbf{R}_{c_i}\textbf{t}_{cb}
    \end{aligned}
    \label{cameraimu}
\end{equation}
where $s$ is the scale, which is set as 1.0 in the paper. The variables that need to be solved are expressed as $x=[\delta b_a\ \ \textbf{g}]^T\in \mathbb{R}^6$. We select three consecutive frames $(i,j,k)$ to construct the residual according to the translation pre-integration in Eq. (\ref{pre-integration}). The inter-frame constraint is as follows:
\begin{equation}
    \begin{aligned}
        &\textbf{R}_{b_i}\Delta \textbf{p}_{b_ib_j}-\textbf{p}_{b_j}+\textbf{p}_{b_i}+\textbf{v}_{b_i} \Delta t_{ij}+\frac{1}{2}\textbf{g}{\Delta t_{ij}}^2\\
        =&\textbf{R}_{b_j}\Delta \textbf{p}_{b_jb_k}-\textbf{p}_{b_k}+\textbf{p}_{b_j}+\textbf{v}_{b_j} \Delta t_{jk}+\frac{1}{2}\textbf{g}{\Delta t_{jk}}^2
    \end{aligned}
    \label{constraint}
\end{equation}
\par
During the solution process we assume that the change of gyroscope bias is zero. From Eq. (\ref{update}), Eq. (\ref{cameraimu}) and Eq. (\ref{constraint}), the residual is written as:
\begin{equation}
    r_j(x)=[A_j\ \ B_j]x-C_j
\end{equation}
where
\begin{equation}
    \begin{aligned}
        A_i&=(\frac{\textbf{R}_{c_i}J^{b_a}_{\Delta \textbf{p}_{b_ib_j}}}{\Delta t_{i,j}}-\frac{\textbf{R}_{c_j}J^{b_a}_{\Delta \textbf{p}_{b_jb_k}}}{\Delta t_{j,k}})\\
        B_i&=-\frac{1}{2}(\Delta t_{j,k}+\Delta t_{i,j})\textbf{I}_{3\times3}\\
        C_i&= (\frac{\textbf{p}_{c_j}-\textbf{p}_{c_i}}{\Delta t_{i,j}}-\frac{\textbf{p}_{c_k}-\textbf{p}_{c_j}}{\Delta t_{j,k}})s+\frac{\textbf{R}_{b_j}\Delta \textbf{p}_{b_jb_k}}{\Delta t_{j,k}}-\frac{\textbf{R}_{b_i}\Delta \textbf{p}_{b_ib_j}}{\Delta t_{i,j}}\\
        &+\textbf{R}_{b_i}\Delta \textbf{v}_{b_ib_j}+(\frac{\textbf{R}_{c_j}-\textbf{R}_{c_i}}{\Delta t_{i,j}}-\frac{\textbf{R}_{c_k}-\textbf{R}_{c_j}}{\Delta t_{j,k}})\textbf{t}_{cb}
    \end{aligned}
\end{equation}
\par
From this, we obtain a constrained optimization problem:
\begin{equation}
\begin{aligned}
    x^*&=\mathop{\arg\min}_{x}||r_p(x)||^2_{\Sigma_p}+\sum_{j=1}^{n}||r_j(x)||^2_{\Sigma_j}\\
    &=x^TDx+Ex+F\\
   ||\textbf{g}||&=G
\end{aligned}
\end{equation}
with
\begin{equation}
    \begin{aligned}
        D&=\Sigma_p^{-1}+\sum_{j=1}^{n}[A_j\ \ B_j]^T\Sigma_j^{-1}[A_j\ \ B_j]\\
   E&=\sum_{j=1}^{n}-2C_j^T\Sigma_j^{-1}[A_j\ \ B_j]\\
   F&=\sum_{j=1}^{n}C_j^T\Sigma_j^{-1}C_j
    \end{aligned}
\end{equation}
where $r_p(x)$ and $\Sigma_p$ are the residual and uncertainty of prior respectively, and $r_k(x)$ and $\Sigma_k$ are the residual errors and their corresponding covariance matrices\cite{inertialonly}. With the constraint that the magnitude of gravity vector $G$ is equal to 9.81, the cost function is solved by Lagrange multiplier method propsed in \cite{analytical}. The analytical solution method does not require any iteration and differentiation, which makes it faster than iterative optimization method. Just like when optimizing the gyroscope bias, and the solved $\delta b_a$ is equal to $b_a$ to be estimated.

\subsubsection{Refinement}
The initial value of the velocity is calculated as follows:
\begin{equation}
    \textbf{v}_{_j}=\frac{\textbf{p}_{_j}-\textbf{p}_{_i}}{\Delta t_{ij}}
\end{equation}
The inertial state variable we have obtained so far is
\begin{equation}
    \mathcal{X}_k=[\textbf{b}\ \ \textbf{g}\ \ \textbf{v}_{0:k}]
\end{equation}
where $\textbf{b}=[b_g\ \ b_a]^T\in \mathbb{R}^6$ is the IMU bias, and $\textbf{g}\in \mathbb{R}^3$ is the gravity vector. $\textbf{v}_{0:k}\in \mathbb{R}^3$ represents the velocities from the first keyframe to the $k^{th}$ keyframe. In the case of obtaining relatively accurate initialization values, we refine all inertial parameters in the local mapping thread \cite{inertialonly}.
\begin{equation}
\begin{aligned}
    \mathcal{X}^*_k&=\mathop{\arg\max}_{\mathcal{X}_k}\ p(\mathcal{X}_k)\prod_{i=1}^kp(\mathcal{I}_{i-1,i}|\textbf{b},\textbf{g}.\textbf{v}_{i-1},\textbf{v}_i)\\
    &=\mathop{\arg\min}_{\mathcal{X}_k}||\textbf{b}||^2_{\Sigma^{-1}_{b}}+\sum^k_{i=1}||r_{\mathcal{I}_{i-1,i}}||^2_{\Sigma^{-1}_{\mathcal{I}_{i-1,i}}}
\end{aligned} 
\end{equation}
\vspace{-0.5cm}
\subsection{Deep Neural Network Integration}
Traditional SLAM systems based on pure geometry and kinematics have limited robustness and accuracy in complex environments due to the lack of high-level understanding of the scene. In order to improve the intelligence of the system, we integrated deep neural networks into it to improve the algorithm's adaptability to complex scenarios. All networks are accelerated by TensorRT to ensure real-time.
\par
To avoid the influence of dynamic objects on pose estimation, we add a new thread namely dynamic feature elimination. In this thread, we combine YOLOv5\cite{yolov5} with DeepSORT\cite{deepsort} to make up for the missed detection problem of YOLOv5. The object tracking network is pre-trained and its execution process runs in parallel with feature extraction and stereo matching. For dynamic objects, the features fall in foreground of their boxes will be eliminated through a method fusing semantic information and geometric constraints\cite{ovd}\cite{cfp}. The detailed steps are shown in Algorithm \ref{algorithm1}.
\par
Traditional feature matching suffers from some issues in case of dim illumination and viewpoint change. Due to the assistance of IMU and to ensure the real-time performance of the system, we only improve the key frame processing process. In this paragraph, we modify the loop closing thread, that is, using DNN-based feature (SuperPoint) and matching method (SuperGlue) instead of traditional feature extraction and matching. When a new keyframe is received, we extract SuperPoint and compute the corresponding bag-of-words in the local mapping thread. When the total number of key frames is greater than 10, we detect the closed-loop frame in the covisibility graph by DBoW2 \cite{ORB-SLAM3}. In the process of pose estimation, SuperGlue makes feature matching in complex environments more accurate and robust. The modified thread provides more closed-loop matching points and more accurate closed-loop pose estimation compared to \cite{ORB-SLAM3} because of the outstanding wide baseline matching capabilities of DNN-based methods.
\begin{algorithm}[!hbp]
\renewcommand{\algorithmicrequire}{\textbf{Require:}}
\caption{The algorithm for dynamic feature elimination}\label{alg:cap}
\begin{algorithmic}[1]
\REQUIRE
\quad
\\ Previous left camera frame $F_{pre}$ and current left camera frame $F_{cur}$; $B \gets$Bounding boxes of $F_{cur}$
\STATE $\textbf{P}_{cur},\textbf{L}_{cur}\gets$ Extract points and lines from $F_{cur}$
\STATE $\textbf{P}_{pre} \gets$ Optical flow tracking
\STATE $F \gets$ Calculate fundamental matrix from $\textbf{P}_{cur},\textbf{P}_{pre}$
\FOR{each pair matching points in $(\textbf{P}_{cur},\textbf{P}_{pre})$}
\STATE $D_i \gets$Calculate epipolar error by epipolar constraints and chi-square distribution
\ENDFOR
\FOR{each box in $B$}
\STATE Sort $D_i$ of the points in the box, and select the epipolar error at the 0.6 position as the static probability $P_{static}$ of the box.
\ENDFOR
\FOR{all boxes whose $P_{static}<0.8$}
\STATE Eliminate points in foreground of the box from $\textbf{P}_{cur}$
\IF{line endpoints are either in foreground}
\STATE Eliminate the line from $\textbf{L}_{cur}$
\ENDIF
\ENDFOR
\STATE \textbf{Output:} Static points $\textbf{P}_{static}$ and lines $\textbf{L}_{static}$
\end{algorithmic}
\label{algorithm1}
\end{algorithm}
\subsection{Visual-Inertial Bundle Adjustment}
In this paper, we define residual errors of visual point observations and inertial measurements as $r_{\mathcal{B}}$  and $r_{\mathcal{C}}^{\mathcal{P}}$ respectively. For the $i^{th}$ and $j^{th}$ frames, the inertial residual error is defined as:
\begin{equation}
    \begin{aligned} {r}_{\mathcal{B}_{b_i,b_j}} & =\left[{r}_{\Delta \textbf{R}_{b_i,b_j}}, {r}_{\Delta \mathrm{v}_{b_i,b_j}}, {r}_{\Delta \mathrm{p}_{b_i,b_j}}\right]\in \mathbb{R}^{9} \\ 
    {r}_{\Delta \mathrm{R}_{b_i,b_j}} & =\mathrm{Log} \left(\Delta \textbf{R}_{b_i,b_j}^{\mathrm{T}} \textbf{R}_{b_i}^{\mathrm{T}} \textbf{R}_{b_{j}}\right) \\ 
    {r}_{\Delta \mathrm{v}_{b_i,b_j}} & =\textbf{R}_{b_i}^{\mathrm{T}}\left(\textbf{v}_{b_j}-\textbf{v}_{b_i}-\textbf{g} \Delta t_{i,j}\right)-\Delta \textbf{v}_{b_i,b_j} \\ 
    {r}_{\Delta \mathrm{p}_{b_i,b_j}} & =\textbf{R}_{b_i}^{\mathrm{T}}\left(\textbf{p}_{b_j}-\textbf{p}_{b_i}-\textbf{v}_{b_i} \Delta t_{i,j}-\frac{1}{2} \textbf{g} \Delta t^{2}_{i,j}\right)-\Delta \textbf{p}_{b_i,b_j}\end{aligned}
\end{equation}
\par
The point re-projection error is stated as:
\begin{equation}
    r_{\mathcal{C}}^{\mathcal{P}_k}={\left\|u_{k}-\pi(RX_k^{w}+t)\right\|}\in \mathbb{R}
\end{equation}
where $u_k$ is a 2D point and $X_k^w$ is the corresponding 3D point in world frane. $\pi(\cdot)$ is the projection function from world frame to image plane. $(R,t)$ is the pose of the frame.
\par
For line segment observations, We introduce two types of errors, namely re-projection error $r_{\mathcal{C}}^{\mathcal{L}_{2D}}$ and back-projection error $r_{\mathcal{C}}^{\mathcal{L}_{3D}}$\cite{plvs}. For the $k^{th}$ line segment, the endpoints in the image plane is $(p_k,q_k)$, and the endpoint in the three-dimensional space is $(P_k,Q_k)$. The re-projection error of line segment is defined as:
\begin{equation}
\begin{aligned}
    r_{\mathcal{C}}^{\mathcal{L}_k^{2D}}&= \left[\begin{array}{c}\left\|d_k \cdot \pi(RP_k^{w}+t)\right\| \\ \left\|d_k \cdot \pi(RQ_k^{w}+t)\right\|\end{array}\right]\in \mathbb{R}^{2}\\
    d_k&=\frac{\overline{p}_k\times \overline{q}_k}{\left\|\overline{p}_k\times \overline{q}_k\right\|}
\end{aligned}
\end{equation}
where $\overline{p}=[p^T,1]^T$, $\overline{q}=[q^T,1]^T$. The back-projection error is written as follows:
\begin{equation}
    \begin{aligned} 
    r_{\mathcal{C}}^{\mathcal{L}_k^{3D}} &=\left[\begin{array}{l}e_{3 D}\left(x_k, {P}^{w}_{k}\right)+\mu e_{P}\left(x_k, {P}^{w}_{k}\right) \\ e_{3 D}\left(x_k, {Q}^{w}_{k}\right)+\mu e_{P}\left(x_k, {Q}^{w}_{k}\right)\end{array}\right] \in \mathbb{R}^{2}\\
    e_{3 D}\left(x_k, {X}^{w}_{k}\right) & = \frac{\left\|\left({X}^{c}_{k}-{\beta}\left({p}_{k}\right)\right) \times\left({X}^{c}_{k}-{\beta}\left({q}_{k}\right)\right)\right\|}{\left\|{\beta}\left({p}_{k}\right)-{\beta}\left({q}_{k}\right)\right\|} \\
    e_{P}\left(x_k, {X}^{w}_{k}\right) & =\left\|{X}^{c}_{k}-{\beta}({x_k})\right\| \\ 
    {X}^{c}_{k} & ={R} {X}^{w}_{k}+{t}
    \end{aligned}
\end{equation}
where $\beta(\cdot)$ represents the back-projection of an endpoint from image plane to camera frame. $\mu$ is a weight used in local BA and is set as 0.5 in this paper. The back-projection error can keep the endpoints of the 3D line segment stable during the optimization process and prevent it from drifting \cite{plvs}. The complete cost function can be expressed as:
\begin{equation}
    \begin{aligned}
        \mathcal{C}=argmin&\sum\| r_{\mathcal{B}_{b_i,b_j}} \|^2_{\Sigma_{B}^{-1}}+\sum_{\mathcal{K}}(\sum_{i\in \mathcal{N}} \rho(\|r_{\mathcal{C}}^{\mathcal{\mathcal{P}}_i}\|_{\Sigma_P^{-1}})+\\
        &\sum_{j\in \mathcal{M}} \rho(\|r_{\mathcal{C}}^{\mathcal{L}_k^{2D}}\|_{\Sigma_{{\mathcal{L}^{2D}}}^{-1}})+\sum_{j\in \mathcal{M}} \rho(\|r_{\mathcal{C}}^{\mathcal{L}_k^{3D}}\|_{\Sigma_{{\mathcal{L}^{3D}}}^{-1}}))
    \end{aligned}
\end{equation}
where $\mathcal{K}$ denotes the set of key frames, and $\mathcal{N}$ and $\mathcal{M}$ represent the number of points and lines in current frame respectively. $\Sigma^{-1}$ represents the corresponding information matrix. $\rho(\cdot)$ is the Huber function, that is defined as $\rho(\|e\|_{\Sigma^{-1}})=e^T\Sigma^{-1}e$.
\section{Experiments and Results}
In this section, we verify the proposed system on public datasets. The public datasets all contain stereo images, IMU measurements and corresponding ground-truth trajectories. All experiments were performed on a computer with Core i7-11700 CPU, 32G RAM and NVIDIA RTX2080 GPU. 
We utilize the Root mean square error (RMSE) of gyroscope bias, accelerometer bias and gravity direction to verify the advancement of our proposed IMU initialization method. Root mean square error (RMSE) of absolute trajectory error (ATE) is used to evaluate the performance of localization. The correct rate represents the tracking completeness in the scene and is used to evaluate the robustness of system. We also analyze the running time of each module of the algorithm to prove its real-time performance.
\subsection{Evaluation of Visual-Inertial Initialization}
\begin{table*}[!htbp]
    \caption{Performance comparison of IMU initialization  in the EuRoC dataset. The top 2 results of each evaluation index are highlighted in \colorbox{gold}{gold} and \colorbox{silver}{silver}.}
    \renewcommand{\arraystretch}{1.0}
    \centering
    \setlength{\tabcolsep}{1.45mm}{
   \begin{tabular}{c|c|*{12}{c}}
    \specialrule{1pt}{\abovetopsep}{\belowbottomsep}
    \multicolumn{2}{c}{\textbf{Sequence}}  & MH01 & MH02 & MH03 & MH04 & MH05 & V101 & V102 & V103 & V201 & V202 & V203 & Avg\\
    \specialrule{0.35pt}{\aboverulesep}{\belowrulesep}
    \multirow{5}{*}{\textbf{Gyro Bias RMSE(\%)}}
    & \textbf{AS-MLE} & 0.39 & 0.17 & 0.21 & 0.20 & 0.23 & 2.21 & 0.89 & 0.89 & 0.80 & 1.25 & 1.10 & 0.76\\
    & \textbf{IO-MAP} & 0.44 & 0.17 & 0.36 & 0.26 & 0.32 & 2.20 & 0.96 & 0.91 & 0.81 & 1.25 & 1.19 & 0.81\\
    & \textbf{DRT-t} & \colorbox{gold}{0.21} & \colorbox{gold}{0.09} & \colorbox{silver}{0.12} & 0.15 & \colorbox{silver}{0.13} & \colorbox{gold}{0.45} & \colorbox{silver}{0.48} & \colorbox{silver}{0.32} & \colorbox{silver}{0.75} & \colorbox{silver}{0.94} & 0.84 & 0.41\\
    & \textbf{DRT-l} & \colorbox{gold}{0.21} & \colorbox{silver}{0.10} & \colorbox{silver}{0.12} & \colorbox{gold}{0.14} & \colorbox{gold}{0.12} & \colorbox{silver}{0.46} & \colorbox{gold}{0.46} & 0.35 & 0.77 & \colorbox{silver}{0.94} & \colorbox{silver}{0.75} & \colorbox{silver}{0.40}\\
    & \textbf{Ours} & 0.26 & 0.12 & \colorbox{gold}{0.11} & \colorbox{gold}{0.14} & 0.16 & \colorbox{silver}{0.46} & 0.51 & \colorbox{gold}{0.17} & \colorbox{gold}{0.58} & \colorbox{gold}{0.66} & \colorbox{gold}{0.71} & \colorbox{gold}{0.35}\\
    \specialrule{0.35pt}{\aboverulesep}{\belowrulesep}
    \multirow{5}{*}{\textbf{Acc Bias RMSE(\%)}}
    & \textbf{AS-MLE} & \colorbox{silver}{82.17} & \colorbox{silver}{28.80} & 30.14 & 39.69 & \colorbox{silver}{14.55} & \colorbox{silver}{20.78} & 25.83 & \colorbox{silver}{31.83} & / & 53.59 & \colorbox{silver}{58.26} & 38.56\\
    & \textbf{IO-MAP} & / & \colorbox{gold}{27.12} & \colorbox{silver}{26.77} & \colorbox{silver}{37.22} & 16.91 & \colorbox{gold}{20.03} & \colorbox{gold}{25.54} & 32.51 & \colorbox{silver}{96.12} & \colorbox{silver}{47.44} & \colorbox{gold}{55.64} & \colorbox{silver}{38.53}\\
    & \textbf{DRT-t} & - & - & - & - & - & - & - & - & - & - & - &\\
    & \textbf{DRT-l} & - & - & - & - & - & - & - & - & - & - & - &\\
    & \textbf{Ours} & \colorbox{gold}{77.99} & 32.30 & \colorbox{gold}{22.56} & \colorbox{gold}{26.62} & \colorbox{gold}{11.21} & 21.07 & \colorbox{silver}{25.77} & \colorbox{gold}{29.68} & \colorbox{gold}{79.83} & \colorbox{gold}{39.16} & 56.52 & \colorbox{gold}{38.43}\\
    \specialrule{0.35pt}{\aboverulesep}{\belowrulesep}
    \multirow{5}{*}{\textbf{G.Dir RMSE(°)}}
    & \textbf{AS-MLE} & \colorbox{silver}{1.27} & 0.41 & \colorbox{silver}{0.44} & 0.49 & 0.23 & \colorbox{silver}{0.93} & \colorbox{gold}{0.36} & \colorbox{silver}{0.46} & 1.57 & 0.70 & \colorbox{silver}{0.80} & \colorbox{silver}{0.70}\\
    & \textbf{IO-MAP} & 11.04 & \colorbox{gold}{0.38} & 0.46 & \colorbox{silver}{0.45} & \colorbox{silver}{0.20} & \colorbox{gold}{0.92} & \colorbox{gold}{0.36} & 0.48 & \colorbox{silver}{1.40} & \colorbox{silver}{0.66} & \colorbox{gold}{0.78} & 1.56\\
    & \textbf{DRT-t} & 2.63 & 2.67 & 3.00 & 3.93 & 2.80 & 3.24 & 3.31 & 3.93 & 2.73 & 2.67 & 2.67 & 3.05\\
    & \textbf{DRT-l} & 2.72 & 2.89 & 2.96 & 3.45 & 2.63 & 3.08 & 2.77 & 2.69 & 2.59 & 2.89 & 3.66 & 2.94\\
    & \textbf{Ours} & \colorbox{gold}{0.95} & \colorbox{silver}{0.39} & \colorbox{gold}{0.26} & \colorbox{gold}{0.33} & \colorbox{gold}{0.19} & 0.94 & 0.38 & \colorbox{gold}{0.41} & \colorbox{gold}{1.02} & \colorbox{gold}{0.62} & 0.81 & \colorbox{gold}{0.58}\\
    \specialrule{1pt}{\aboverulesep}{\belowrulesep}
    \end{tabular}}
    \label{imu}
\end{table*}
The EuRoC\cite{euroc} dataset contains stereo images and corresponding IMU measurements collected by a micro aerial vehicle and is widely used in evaluation of visual-inertial systems. There are eleven sequences in total, divided into three difficulty levels: easy(MH01/2, V101,V201), medium(MH03/4, V102, V202) and difficult(MH05, V103, V203). The groundtruth is collected by vicon motion capture system, including pose and IMU bias at each moment.
\par
To evaluate the IMU initialization performance of PLE-SLAM, we compared it against AS-MLE\cite{asmle} and DRT\cite{he2023rotation} (including DRT-t and DRT-l). As shown in Table \ref{imu}, the proposed approach in this paper outperforms other methods on almost all sequences due to the introduction of line features. Because DRT-t and DRT-l assume the accelerometer bias is zero, the corresponding error is not given in the results. We use / to indicate the error that exceeds 100\%. The gravity direction error of DRTs are all more than 2.5 degree due to the ignorance of accelerometer bias, while the gravity direction estimation error of other algorithms that take the accelerometer bias into account is almost no more than 1 degree.
\begin{table*}[!htbp]
    \caption{Performance comparison of localization accuracy in the EuRoC dataset (RMS ATE in m.). The top 3 results of each column are highlighted in \colorbox{gold}{gold}, \colorbox{silver}{silver} and \colorbox{bronze}{bronze}.}
    \renewcommand{\arraystretch}{1.05}
    \centering
    \setlength{\tabcolsep}{1.7mm}{
    \begin{tabular}{c|c|*{12}{c}}
        \specialrule{1pt}{\abovetopsep}{\belowbottomsep}
         \multicolumn{2}{c}{\textbf{Sequence}}  & MH01 & MH02 & MH03 & MH04 & MH05 & V101 & V102 & V103 & V201 & V202 & V203 & Avg\\
        \specialrule{0.35pt}{\aboverulesep}{\belowrulesep}
        \multirow{3}{*}{\textbf{VIO}}
        & \textbf{DM-VIO}             & 0.065  & 0.044  & 0.097  & 0.102  & 0.096  & 0.048  & 0.045  & 0.069  & 0.029  & 0.050  & 0.114  & 0.069  \\
        & \textbf{RD-VIO}             & 0.109  & 0.115  & 0.141  & 0.247  & 0.267  & 0.060  & 0.091  & 0.133  & 0.058  & 0.100  & 0.147  & 0.133  \\
        & \textbf{EPLF-VINS}      & 0.043  & 0.072  & 0.047  & 0.125  & 0.085  & 0.046  & 0.034  & 0.123  & 0.096  & 0.058  & 0.111  & 0.076  \\
        \specialrule{0.35pt}{\aboverulesep}{\belowrulesep}
        \multirow{6}{*}{\textbf{SLAM}}
        & \textbf{ORB-SLAM3}  & 0.036  & 0.033  & 0.035  & \colorbox{silver}{0.051}  & 0.082  & \colorbox{bronze}{0.038}  & \colorbox{silver}{0.014}  & 0.024  & 0.032  & \colorbox{silver}{0.014}  & \colorbox{bronze}{0.024}  & 0.035  \\
        & \textbf{OKVIS2}            & \colorbox{silver}{0.027}  & \colorbox{silver}{0.023}  & \colorbox{silver}{0.028}  & 0.066  & \colorbox{silver}{0.068}  & \colorbox{silver}{0.035}  & \colorbox{gold}{0.013}  & \colorbox{gold}{0.019}  & \colorbox{silver}{0.023}  & \colorbox{bronze}{0.015}  & \colorbox{silver}{0.020}  & \colorbox{silver}{0.031}  \\
        & \textbf{Dynam-SLAM} & 0.078  & 0.067  & 0.061  & 0.077  & 0.096  & 0.052  & 0.040  & 0.051  & 0.048  & 0.056  & 0.085  & 0.065  \\
        & \textbf{DVI-SLAM}        & 0.042  & 0.046  & 0.081  & 0.072  & 0.069  & 0.059  & 0.034  & 0.028  & 0.040  & 0.039  & 0.055  & 0.051  \\
        & \textbf{MAVIS}               & \colorbox{gold}{0.024}  & \colorbox{bronze}{0.025}  & \colorbox{bronze}{0.032}  & \colorbox{bronze}{0.053}  & \colorbox{bronze}{0.075}  & \colorbox{gold}{0.034}  & 0.016  & \colorbox{silver}{0.021}  & \colorbox{bronze}{0.031}  & 0.021  & 0.039  & \colorbox{bronze}{0.034}  \\
        & \textbf{PLE-SLAM(ours)}              & \colorbox{bronze}{0.029}  & \colorbox{gold}{0.016}  & \colorbox{gold}{0.027}  &\colorbox{gold}{0.043}  & \colorbox{gold}{0.047}  & 0.039  & \colorbox{silver}{0.014}  & \colorbox{bronze}{0.022}  & \colorbox{gold}{0.015}  & \colorbox{gold}{0.013}  & \colorbox{gold}{0.019}  & \colorbox{gold}{0.026} \\
        \specialrule{1pt}{\aboverulesep}{\belowrulesep}
    \end{tabular}}
    \label{ateoneuroc}
\end{table*}
\subsection{Evaluation of Localization Accuracy}
\subsubsection{EuRoC Dataset}
To further evaluate the localization accuracy on EuRoC dataset, we selected DM-VIO\cite{dmvio},  RD-VIO\cite{rdvio}, EPLF-VINS\cite{eplfvins}, ORB-SLAM3\cite{ORB-SLAM3}, OKVIS2\cite{okvis2}, Dynam-SLAM\cite{dynamslam}, DVI-SLAM\cite{dvislam} and MAVIS\cite{mavis} for comparative experiments. In Table \ref{ateoneuroc}, we use absolute trajectory error (ATE) to evaluate the localization accuracy of these methods. PLE-SLAM achieved best performance on 7 of 11 sequences. Among all SLAM algorithms, only OKVIS2 and MAVIS achieved slightly lower accuracy than our method. Compared to ORB-SLAM3, PLE-SLAM achieve more than 20\% performance improvements on most sequences, especially medium and difficult sequences.
\subsubsection{OpenLORIS-Scene Dataset}
\begin{figure*}[!htbp]
\centerline{\includegraphics[width=\textwidth]{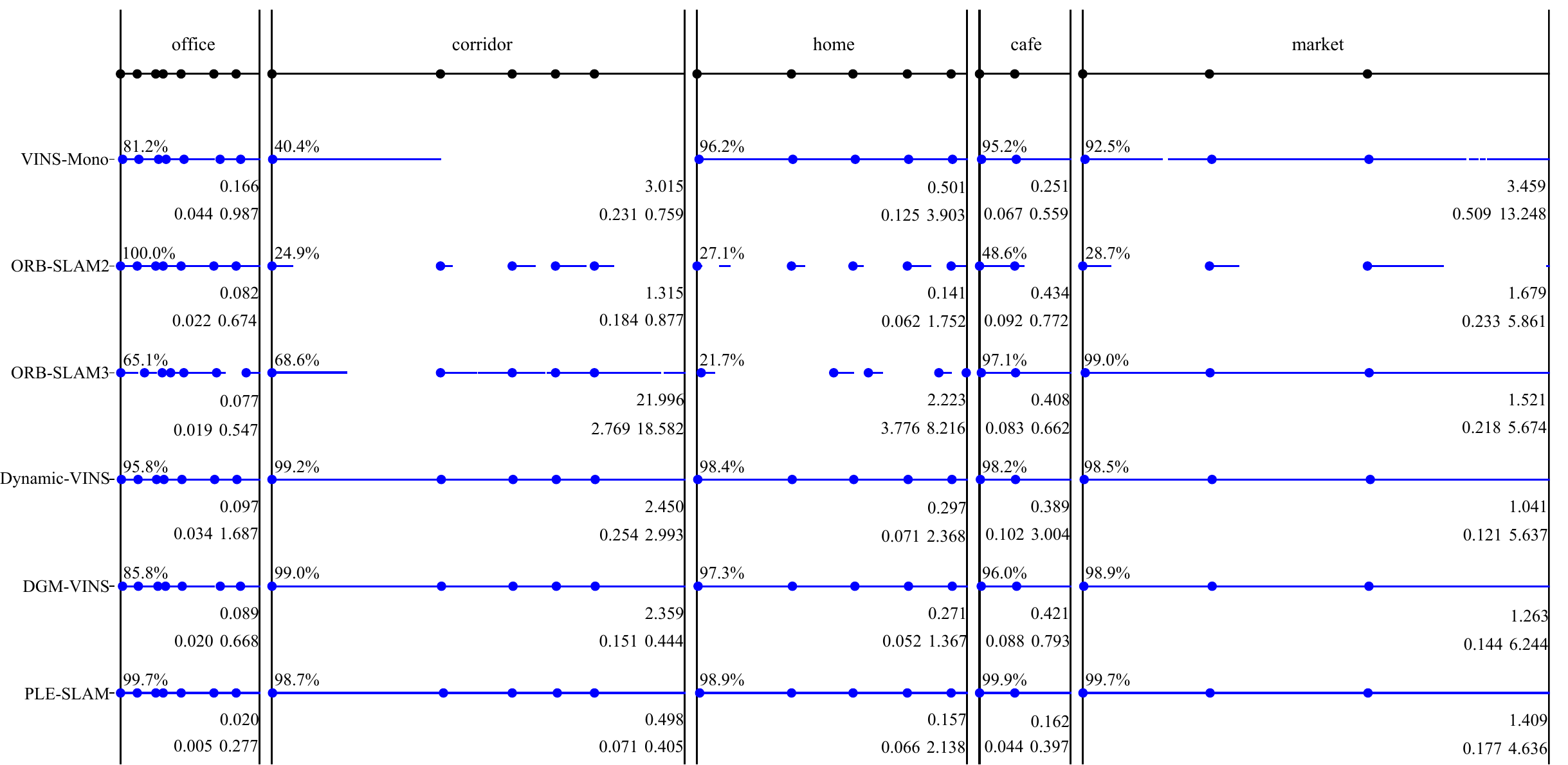}}
\caption{Per-sequence testing results with the OpenLORIS-Scene datasets. Each black dot on the top line represents the start of one data sequence. For each algorithm, blue dots indicate successful initialization, and blue lines indicate successful tracking. The percentage value on the top left of each scene is average correct rate, larger means more robust. The float value in the first line on the bottom right is average ATE RMSE and the two values in the second line from left to right are average T.RPE and average R.RPE respectively, smaller means more accurate. Parts with errors that are larger than excessive are directly ignored during the entire evaluation process and their corresponding correct rate parts are eliminated.}
\label{openloris}
\end{figure*}

\begin{figure}[!bp]
\centerline{\includegraphics[width=\columnwidth]{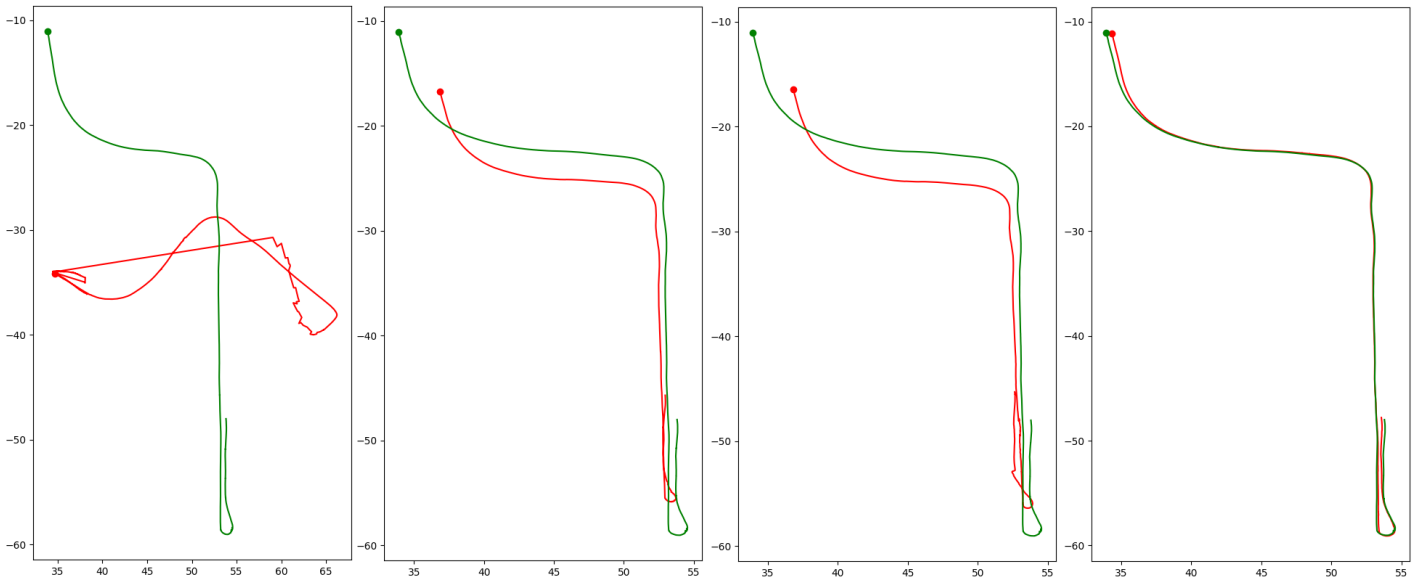}}
\caption{Trajectory comparison results on corridor1-3. Green line denotes the grountruth and red line denotes the estimated trajectory. From left to right in the figure are ORB-SLAM3, Dynamic-VINS, DGM-VINS and PLE-SLAM.}
\label{corridor}
\end{figure}

The OpenLORIS-Scene\cite{openloris} dataset is collected in real life environment and contains various challenging scenes like dim illumination, dynamic object interference, viewpoint change and low-texture areas, which brings huge difficulties to the SLAM algorithm. The testing results are shown in Fig. \ref{openloris}, including VINS-Mono\cite{VINS-Mono}, ORB-SLAM2/3\cite{ORB-SLAM2,ORB-SLAM3}, Dynamic-VINS\cite{dynamicvins} and DGM-VINS\cite{dgmvins}. Visual-inertial systems is significantly better than the pure visual system in terms of correct rate. Although Dynamic-VINS and DGM-VINS are optimized for complex scenes, their positioning accuracy is still not as good as our proposed method. Point-line feature fusion and efficient IMU initialization enable our system to achieve the best performance on the entire dataset. A detailed trajectory comparison on a certain sequence is shown in the Fig. \ref{corridor}.
\begin{table}[htbp]
    \caption{Performance comparison in the TUM-VI dataset (RMS ATE in m.). The top 2 results of each row are highlighted in \colorbox{gold}{gold} and \colorbox{silver}{silver}.}
    \renewcommand{\arraystretch}{1.0}
    \centering
    \setlength{\tabcolsep}{0.55mm}{
    \begin{threeparttable}
    \begin{tabular}{*{7}{c}}
    \specialrule{1pt}{\aboverulesep}{\belowrulesep}
    Sequence        & ORB-SLAM3     & DM-VIO    & OKVIS2    & PLE-SLAM      & Length[m]     &LC\\
    \specialrule{0.35pt}{\aboverulesep}{\belowrulesep}
    corridor1       & \colorbox{silver}{0.03}                      & 0.19           & \colorbox{gold}{0.02}         & 0.06  & 305  &\checkmark\\
    corridor2       & \colorbox{gold}{0.02}                      & 0.47           & 0.06         & \colorbox{gold}{0.02}  & 322  &\checkmark\\
    corridor3       & \colorbox{gold}{0.02}                      & 0.24           & 0.03         & \colorbox{gold}{0.02}  & 300  &\checkmark\\
    corridor4       & 0.21                      & 0.13           & \colorbox{silver}{0.10}         & \colorbox{gold}{0.08}  & 114  &\\ 
    corridor5       & \colorbox{gold}{0.01}                      & 0.16           & 0.09         & \colorbox{silver}{0.02}  & 270  &\checkmark\\ 
    magistrale1 & 0.24                      &  2.35           & \colorbox{gold}{0.07}        & \colorbox{silver}{0.19}  & 918  &\checkmark\\ 
    magistrale2 & \colorbox{silver}{0.52}                      & 2.24            & 1.22        & \colorbox{gold}{0.12}  & 561  &\checkmark\\ 
    magistrale3 & 1.86                      & 1.69            & \colorbox{gold}{0.09}        & \colorbox{silver}{0.49}  & 566  &(\checkmark)\\ 
    magistrale4 & \colorbox{silver}{0.16}                      & 1.02            & 0.25        & \colorbox{gold}{0.15}  & 688  &\checkmark\\ 
    magistrale5 & 1.13                      & 0.73            & \colorbox{gold}{0.02}        & \colorbox{silver}{0.34}  & 458  &\checkmark\\ 
    magistrale6 & 0.97                      & 1.19            & \colorbox{silver}{0.76}        & \colorbox{gold}{0.22}  & 771  &(\checkmark)\\ 
    slides1          & 0.41                     & \colorbox{silver}{0.31}          & 0.37         & \colorbox{gold}{0.15}  & 289  &(\checkmark)\\ 
    slides2          & 0.49                     & 0.87          & \colorbox{silver}{0.16}         & \colorbox{gold}{0.09}  & 299  &(\checkmark)\\ 
    slides3          & 0.47                     & 0.60          & \colorbox{gold}{0.13}         & \colorbox{silver}{0.25}  & 383 &\\ 
    \specialrule{1pt}{\aboverulesep}{\belowrulesep}
    \end{tabular}
    \label{tumvi}
         1. The results of the comparison algorithms are all from their corresponding papers.
         
         2. (\checkmark) represents the sequence where loops are only detected by PLE-SLAM compared to ORB-SLAM3
    \end{threeparttable}}
\end{table}
\subsubsection{TUM-VI Dataset}
The TUM-VI\cite{tumvi} dataset is collected by a hand-held fisheye stereo-inertial device. The groundtruth is obtained through a mothion capture system. We select four scenes(corridor, magistrale, romm and slides) that contains challenging motion modes and places, such as rapid rotation, low-texture, dynamic illumination and large scale scene, etc. We compared our system against three recent excellent visual-inertial systems, among which DM-VIO\cite{dmvio} is an odometry without loop closure, ORB-SLAM3\cite{ORB-SLAM3} and OKVIS2\cite{okvis2} are full SLAM systems. 
\par
All algorithms except DM-VIO achieve localization accuracy of 0.01m in terms of RMS ATE on the room sequences. Other results are shown in Table \ref{tumvi} and demonstrate the superiority of the proposed method. The modified loop closing thread enables our system to detect more loops and achieve more accurate positioning than ORB-SLAM3. The performance of OKVIS2 is similar to our algorithm, and is sightly inferior in some sequences, like magistrale2/4/6 and slides1/2.
\subsection{Ablation Study}
To show the effectiveness of each proposed module, we removed them individually. We use L to represent the introduction of line, E to represent the proposed IMU initialization, and D to represent the integration of DNNs. The comparison results are shown in the Table \ref{ablation}. Some modes have errors exceeding 10m or are unable to complete all tracking on some sequences, we use $-$ to indicate them. There is a long dim corridor in corridor1-3 and our method provides more accurate inertial parameters and line constraints which avoids excessive IMU drift in the case of insufficient points. The introduction of line features provides more effective visual constraints in scenes that can be tracked normally and leads to better performance. In dynamic environments, DNNs remove the dynamic features and provide more efficient and accurate loop closure(magistrale6, slides2 and market1-3).  In summary, the method with the improved module has a certain accuracy improvement compared to the baseline and the full PLE-SLAM has the best performance on almost all sequences.
\begin{table}[!htbp]
    \caption{Results of ablation experiments in terms of localization accuracy in the above public datasets (RMS ATE in m.). The best result or each column is highlighted in \colorbox{gold}{gold}.}
    \renewcommand{\arraystretch}{1.0}
    \centering
    \setlength{\tabcolsep}{0.55mm}{
    \begin{tabular}{c|*{7}{c}}
        \specialrule{1pt}{\abovetopsep}{\belowbottomsep}
         \multicolumn{1}{c}{\textbf{Sequence}}  & \begin{tabular}[c]{@{}c@{}}MH05\\difficult\end{tabular} & \begin{tabular}[c]{@{}c@{}}V103\\difficult\end{tabular} & \begin{tabular}[c]{@{}c@{}}V203\\difficult\end{tabular} & \begin{tabular}[c]{@{}c@{}}magistrale\\6\end{tabular} & \begin{tabular}[c]{@{}c@{}}slides\\2\end{tabular} & \begin{tabular}[c]{@{}c@{}}corridor\\1-3\end{tabular} & \begin{tabular}[c]{@{}c@{}}market\\1-3\end{tabular}\\
        \specialrule{0.35pt}{\aboverulesep}{\belowrulesep}
        \textbf{Baseline}     & 0.077 & 0.026 & 0.032 & 0.836 & 0.463 & — & 1.587\\
        \textbf{Baseline+L}        & 0.052 & \colorbox{gold}{0.022} & 0.024 & 0.753 & — & 2.591 & 1.605 \\
        \textbf{Baseline+E}       & 0.066 & 0.025 & 0.020 & 0.623 & 0.317 & 0.443 & 1.362\\
        \textbf{Baseline+D}       & 0.075 & 0.028 & 0.031 & \colorbox{gold}{0.219} & 0.142 & — & 1.122\\
        \textbf{PLE-SLAM} & \colorbox{gold}{0.047} & \colorbox{gold}{0.022} & \colorbox{gold}{0.019} & 0.223 & \colorbox{gold}{0.093} & \colorbox{gold}{0.205} & \colorbox{gold}{0.947}\\
        \specialrule{1pt}{\aboverulesep}{\belowrulesep}
    \end{tabular}}
    \label{ablation}
\end{table}
\subsection{Runtime Analysis}
Table \ref{inertialtime} shows the time comparison results of IMU initialization in the V2 sequences of EuRoC dataset. Our method achieves more accurate inertial parameters estimation without obvious disadvantages in terms of time consumption. Accurate initial values estimation makes our algorithm less time-consuming during refinement compared to IO-MAP. Compared with DRT-t and DRT-l, our algorithm introduces additional line constraints when directly solving rotation to optimize gyroscope bias. Although it improves the progress, it also increases the amount of calculation. The average running time of each module is shown in Table \ref{time}. Our algorithm can achieve an average tracking speed of nearly 47 FPS, fully meeting real-time requirements.
\begin{table}[!htbp]
    \caption{Processing time of IMU initialization (Mean time in ms).}
    \renewcommand{\arraystretch}{1.0}
    \centering
    \setlength{\tabcolsep}{1.2mm}{
    \begin{tabular}{c*{5}{c}}
        \specialrule{1pt}{\abovetopsep}{\belowbottomsep}
         \multicolumn{1}{c}{\textbf{Module}}  & IO-MAP & AS-MLE & DRT-t & DRT-l & Ours\\
        \specialrule{0.35pt}{\aboverulesep}{\belowrulesep}
        \textbf{Bg Est.}     & \multirow{3}{*}{{12.6}}  & 0.11 &1.77 & 1.78 & 2.11\\
        \textbf{Ba \&/or Grav Est.} & & 0.06 & 2.57 & 1.33 & 0.05\\
        \textbf{Refinement} & & — & — & — & 0.89\\
        \specialrule{0.35pt}{\aboverulesep}{\belowrulesep}
        \multicolumn{1}{c}{\textbf{Total Cost}}  & 12.6 & 0.17 & 4.34 & 3.11 & 3.05 \\
        \specialrule{1pt}{\aboverulesep}{\belowrulesep}
    \end{tabular}}
    \label{inertialtime}
\end{table}
\begin{table*}[!htbp]
    \caption{Running time of each module (Mean time in ms).}
    \renewcommand{\arraystretch}{1.0}
    \centering
    \setlength{\tabcolsep}{0.56mm}{
    \begin{tabular}{c*{11}{c}}
        \specialrule{1pt}{\abovetopsep}{\belowbottomsep}
         \multicolumn{1}{c}{\textbf{Module}}  & \begin{tabular}[c]{@{}c@{}}Point \& Line\\Extraction\end{tabular} & \begin{tabular}[c]{@{}c@{}}Stereo\\Matching\end{tabular} & \begin{tabular}[c]{@{}c@{}}Object\\Tracking\end{tabular} & \begin{tabular}[c]{@{}c@{}}Dynamic Feature\\Elimination\end{tabular} & \begin{tabular}[c]{@{}c@{}}IMU\\Preintergration\end{tabular} & \begin{tabular}[c]{@{}c@{}}Pose\\Optimization\end{tabular} & \begin{tabular}[c]{@{}c@{}}Tracking \\Per Frame\end{tabular} & \begin{tabular}[c]{@{}c@{}}SuperPoint\\Extraction\end{tabular} & \begin{tabular}[c]{@{}c@{}}Local\\BA\end{tabular} & \begin{tabular}[c]{@{}c@{}}SuperGlue\\Matching\end{tabular} & \begin{tabular}[c]{@{}c@{}}Full\\BA\end{tabular}\\
        \specialrule{0.35pt}{\aboverulesep}{\belowrulesep}
        \begin{tabular}[c]{@{}c@{}}\textbf{Running}\\\textbf{Time}\end{tabular}     & 12.58 & 1.94 & 14.41 & 1.07 & 0.13 & 1.78 & 21.50 & 5.24 & 71.52 & 10.15& \begin{tabular}[c]{@{}c@{}}647.96\\(127 key frames)\end{tabular} \\
        \specialrule{1pt}{\aboverulesep}{\belowrulesep}
    \end{tabular}}
    \label{time}
\end{table*}
\section{Conclusion}
We propose an accurate stereo visual-inertial SLAM system based on point-line features. The method extract line segments in parallel with point extraction, and multi-threading technology is used to accelerate the processing of pyramid images and feature grids. In order to maintain the stability of line segments during tracking, we merge short line segments according to their their inclinations and vertical distances. After fusing, the lines that is shorter than pre-defined threshold will be filtered out. The introduction of line improves tracking stability and accuracy compared to point-only method. We also propose an efficient visual-inertial initialization method, independently estimating gyroscope bias and accelerometer bias in two steps. The gyroscope bias is directly optimized using raw point-line observations and inertial measurements without SfM. The accelerometer bias and the gravity direction are solved by an analytical solution which runs faster than iterative solution. All initial state variables will be optimized in back-end through MAP. We also integrated multiple neural networks into the system to obtain semantic information and process key frames, significantly improving the robustness and localization accuracy in challenging environments. The experimental results on public datasets and real world verify the effectiveness of our proposed algorithm and illustrate that our method is one of the state-of-the-art visual-inertial SLAM systems.




 
%
\bibliographystyle{IEEEtran}
\bibliography{IEEEabrv,reference}

\begin{thebibliography}{10}
\providecommand{\url}[1]{#1}
\csname url@samestyle\endcsname
\providecommand{\newblock}{\relax}
\providecommand{\bibinfo}[2]{#2}
\providecommand{\BIBentrySTDinterwordspacing}{\spaceskip=0pt\relax}
\providecommand{\BIBentryALTinterwordstretchfactor}{4}
\providecommand{\BIBentryALTinterwordspacing}{\spaceskip=\fontdimen2\font plus
\BIBentryALTinterwordstretchfactor\fontdimen3\font minus \fontdimen4\font\relax}
\providecommand{\BIBforeignlanguage}[2]{{%
\expandafter\ifx\csname l@#1\endcsname\relax
\typeout{** WARNING: IEEEtran.bst: No hyphenation pattern has been}%
\typeout{** loaded for the language `#1'. Using the pattern for}%
\typeout{** the default language instead.}%
\else
\language=\csname l@#1\endcsname
\fi
#2}}
\providecommand{\BIBdecl}{\relax}
\BIBdecl

\bibitem{SVO}
C.~Forster, M.~Pizzoli, and D.~Scaramuzza, ``{SVO}: Fast semi-direct monocular visual odometry,'' in \emph{Proc. IEEE Int. Conf. Robot Autom. (ICRA)}.\hskip 1em plus 0.5em minus 0.4em\relax IEEE, 2014, pp. 15--22.

\bibitem{Forster17troSVO}
C.~Forster, Z.~Zhang, M.~Gassner, M.~Werlberger, and D.~Scaramuzza, ``{SVO}: Semidirect visual odometry for monocular and multicamera systems,'' \emph{{IEEE} Trans. Robot.}, vol.~33, no.~2, pp. 249--265, 2017.

\bibitem{ORB-SLAM2}
R.~Mur-Artal and J.~D. Tard{\'o}s, ``{ORB-SLAM2: An} open-source {SLAM} system for monocular, stereo, and {RGB-D} cameras,'' \emph{{IEEE} Trans. Robot.}, vol.~33, no.~5, pp. 1255--1262, 2017.

\bibitem{DSO}
J.~Engel, V.~Koltun, and D.~Cremers, ``Direct sparse odometry,'' \emph{{IEEE} Trans. Pattern Anal. Mach. Intell.}, vol.~40, no.~3, pp. 611--625, 2018.

\bibitem{OKVIS}
S.~Leutenegger, S.~Lynen, M.~Bosse, R.~Siegwart, and P.~Furgale, ``Keyframe-based visual--inertial odometry using nonlinear optimization,'' \emph{Int. J. Robot. Res.}, vol.~34, no.~3, pp. 314--334, 2015.

\bibitem{VINS-Mono}
T.~Qin, P.~Li, and S.~Shen, ``{VINS-Mono}: A robust and versatile monocular visual-inertial state estimator,'' \emph{{IEEE} Trans. Robot.}, vol.~34, no.~4, pp. 1004--1020, 2018.

\bibitem{VI-ORB-SLAM}
R.~Mur-Artal and J.~D. Tard{\'o}s, ``Visual-inertial monocular {SLAM} with map reuse,'' \emph{IEEE Robot. Autom. Lett.}, vol.~2, no.~2, pp. 796--803, 2017.

\bibitem{vidso}
L.~Von~Stumberg, V.~Usenko, and D.~Cremers, ``Direct sparse visual-inertial odometry using dynamic marginalization,'' in \emph{Proc. IEEE Int. Conf. Robot Autom. (ICRA)}.\hskip 1em plus 0.5em minus 0.4em\relax IEEE, 2018, pp. 2510--2517.

\bibitem{SVOPro}
D.~S. Giovanni~Cioffi, ``Tightly-coupled fusion of global positional measurements in optimization-based visual-inertial odometry,'' in \emph{Proc. IEEE Int. Conf. Intel. Robots and Syst. (IROS)}, 2020.

\bibitem{ORB-SLAM3}
C.~Campos, R.~Elvira, J.~J.~G. Rodr{\'\i}guez, J.~M. Montiel, and J.~D. Tard{\'o}s, ``{ORB-SLAM3: An} accurate open-source library for visual, visual--inertial, and multimap slam,'' \emph{{IEEE} Trans. Robot.}, vol.~37, no.~6, pp. 1874--1890, 2021.

\bibitem{orb}
E.~Rublee, V.~Rabaud, K.~Konolige, and G.~Bradski, ``{ORB: An} efficient alternative to {SIFT} or {SURF},'' in \emph{Proc. IEEE Int. Conf. Comput. Vis. (ICCV)}.\hskip 1em plus 0.5em minus 0.4em\relax Ieee, 2011, pp. 2564--2571.

\bibitem{plsvo}
R.~Gomez-Ojeda, J.~Briales, and J.~Gonzalez-Jimenez, ``{PL-SVO}: Semi-direct monocular visual odometry by combining points and line segments,'' in \emph{Proc. IEEE Int. Conf. Intel. Robots and Syst. (IROS)}.\hskip 1em plus 0.5em minus 0.4em\relax IEEE, 2016, pp. 4211--4216.

\bibitem{plslammono}
A.~Pumarola, A.~Vakhitov, A.~Agudo, A.~Sanfeliu, and F.~Moreno-Noguer, ``{PL-SLAM}: Real-time monocular visual {SLAM} with points and lines,'' in \emph{Proc. IEEE Int. Conf. Robot Autom. (ICRA)}.\hskip 1em plus 0.5em minus 0.4em\relax IEEE, 2017, pp. 4503--4508.

\bibitem{plslam}
R.~Gomez-Ojeda, F.-A. Moreno, D.~Zuniga-No{\"e}l, D.~Scaramuzza, and J.~Gonzalez-Jimenez, ``{PL-SLAM}: A stereo {SLAM} system through the combination of points and line segments,'' \emph{IEEE Trans. Robot.}, vol.~35, no.~3, pp. 734--746, 2019.

\bibitem{airvo}
K.~Xu, Y.~Hao, S.~Yuan, C.~Wang, and L.~Xie, ``Airvo: An illumination-robust point-line visual odometry,'' in \emph{Proc. IEEE Int. Conf. Intel. Robots and Syst. (IROS)}.\hskip 1em plus 0.5em minus 0.4em\relax IEEE, 2023, pp. 3429--3436.

\bibitem{plvio}
Y.~He, J.~Zhao, Y.~Guo, W.~He, and K.~Yuan, ``{PL-VIO}: Tightly-coupled monocular visual--inertial odometry using point and line features,'' \emph{Sensors}, vol.~18, no.~4, p. 1159, 2018.

\bibitem{pl-vins}
Q.~Fu, J.~Wang, H.~Yu, I.~Ali, F.~Guo, Y.~He, and H.~Zhang, ``{PL-VINS}: Real-time monocular visual-inertial {SLAM} with point and line features,'' 2020.

\bibitem{tvtplslam}
X.~Liu, S.~Wen, and H.~Zhang, ``A real-time stereo visual-inertial {SLAM} system based on point-and-line features,'' \emph{IEEE Trans. Veh.}, 2023.

\bibitem{eplfvins}
L.~Xu, H.~Yin, T.~Shi, D.~Jiang, and B.~Huang, ``{EPLF-VINS}: Real-time monocular visual-inertial {SLAM} with efficient point-line flow features,'' \emph{IEEE Robot. Autom. Lett.}, vol.~8, no.~2, pp. 752--759, 2022.

\bibitem{edlines}
C.~Akinlar and C.~Topal, ``{EDLines}: A real-time line segment detector with a false detection control,'' \emph{Pattern Recognit Lett}, vol.~32, no.~13, pp. 1633--1642, 2011.

\bibitem{tightclosed}
A.~Martinelli, ``Closed-form solution of visual-inertial structure from motion,'' \emph{Int J Comput Vis}, vol. 106, no.~2, pp. 138--152, 2014.

\bibitem{openvins}
P.~Geneva, K.~Eckenhoff, W.~Lee, Y.~Yang, and G.~Huang, ``{OpenVINS}: A research platform for visual-inertial estimation,'' in \emph{Proc. IEEE Int. Conf. Robot Autom. (ICRA)}.\hskip 1em plus 0.5em minus 0.4em\relax IEEE, 2020, pp. 4666--4672.

\bibitem{vinsintiliaztion}
T.~Qin and S.~Shen, ``Robust initialization of monocular visual-inertial estimation on aerial robots,'' in \emph{Proc. IEEE Int. Conf. Intel. Robots and Syst. (IROS)}.\hskip 1em plus 0.5em minus 0.4em\relax IEEE, 2017, pp. 4225--4232.

\bibitem{inertialonly}
C.~Campos, J.~M. Montiel, and J.~D. Tard{\'o}s, ``Inertial-only optimization for visual-inertial initialization,'' in \emph{Proc. IEEE Int. Conf. Robot Autom. (ICRA)}.\hskip 1em plus 0.5em minus 0.4em\relax IEEE, 2020, pp. 51--57.

\bibitem{analytical}
D.~Zu{\~n}iga-No{\"e}l, F.-A. Moreno, and J.~Gonzalez-Jimenez, ``An analytical solution to the {IMU} initialization problem for visual-inertial systems,'' \emph{IEEE Robot. Autom. Lett.}, vol.~6, no.~3, pp. 6116--6122, 2021.

\bibitem{drt}
Y.~He, B.~Xu, Z.~Ouyang, and H.~Li, ``A rotation-translation-decoupled solution for robust and efficient visual-inertial initialization,'' in \emph{Proc. IEEE Conf. Comput. Vis. Pattern Recognit. (CVPR)}, 2023, pp. 739--748.

\bibitem{efficient}
L.~Kneip and H.~Li, ``Efficient computation of relative pose for multi-camera systems,'' in \emph{Proc. IEEE Conf. Comput. Vis. Pattern Recognit. (CVPR)}, 2014, pp. 446--453.

\bibitem{superpoint}
D.~DeTone, T.~Malisiewicz, and A.~Rabinovich, ``Superpoint: Self-supervised interest point detection and description,'' in \emph{Proc. IEEE Conf. Comput. Vis. Pattern Recognit. (CVPR)}, 2018, pp. 224--236.

\bibitem{superglue}
P.-E. Sarlin, D.~DeTone, T.~Malisiewicz, and A.~Rabinovich, ``Superglue: Learning feature matching with graph neural networks,'' in \emph{Proc. IEEE Conf. Comput. Vis. Pattern Recognit. (CVPR)}, 2020, pp. 4938--4947.

\bibitem{lsd}
R.~G. Von~Gioi, J.~Jakubowicz, J.-M. Morel, and G.~Randall, ``{LSD}: A line segment detector,'' \emph{Image Process. Line}, vol.~2, pp. 35--55, 2012.

\bibitem{fast}
H.~Wei, T.~Zhang, and L.~Zhang, ``A fast analytical two-stage initial-parameters estimation method for monocular-inertial navigation,'' \emph{IEEE Transactions on Instrumentation and Measurement}, vol.~71, pp. 1--12, 2022.

\bibitem{okvis2}
S.~Leutenegger, ``{OKVIS2}: Realtime scalable visual-inertial {SLAM} with loop closure,'' \emph{arXiv preprint arXiv:2202.09199}, 2022.

\bibitem{dynamicvins}
J.~Liu, X.~Li, Y.~Liu, and H.~Chen, ``{RGB-D} inertial odometry for a resource-restricted robot in dynamic environments,'' \emph{IEEE Robot. Autom. Lett.}, vol.~7, no.~4, pp. 9573--9580, 2022.

\bibitem{dgmvins}
B.~Song, X.~Yuan, Z.~Ying, B.~Yang, Y.~Song, and F.~Zhou, ``{DGM-VINS}: Visual-inertial {SLAM} for complex dynamic environments with joint geometry feature extraction and multiple object tracking,'' \emph{IEEE Trans Instrum Meas}, 2023.

\bibitem{ovd}
J.~He, M.~Li, Y.~Wang, and H.~Wang, ``{OVD-SLAM}: An online visual {SLAM} for dynamic environments,'' \emph{IEEE Sens. J}, 2023.

\bibitem{cfp}
X.~Hu, Y.~Zhang, Z.~Cao, R.~Ma, Y.~Wu, Z.~Deng, and W.~Sun, ``{CFP-SLAM}: A real-time visual {SLAM} based on coarse-to-fine probability in dynamic environments,'' in \emph{Proc. IEEE Int. Conf. Intel. Robots and Syst. (IROS)}.\hskip 1em plus 0.5em minus 0.4em\relax IEEE, 2022, pp. 4399--4406.

\bibitem{gcnv2}
J.~Tang, L.~Ericson, J.~Folkesson, and P.~Jensfelt, ``Gcnv2: Efficient correspondence prediction for real-time slam,'' \emph{IEEE Robot. Autom. Lett.}, vol.~4, no.~4, pp. 3505--3512, 2019.

\bibitem{srvio}
A.~Samadzadeh and A.~Nickabadi, ``Srvio: Super robust visual inertial odometry for dynamic environments and challenging loop-closure conditions,'' \emph{IEEE Trans. Robot.}, 2023.

\bibitem{manifold}
C.~Forster, L.~Carlone, F.~Dellaert, and D.~Scaramuzza, ``On-manifold preintegration for real-time visual-inertial odometry,'' \emph{IEEE Trans. Robot.}, vol.~33, no.~1, pp. 1--21, 2016.

\bibitem{he2023rotation}
Y.~He, B.~Xu, Z.~Ouyang, and H.~Li, ``A rotation-translation-decoupled solution for robust and efficient visual-inertial initialization,'' in \emph{Proc. IEEE Conf. Comput. Vis. Pattern Recognit. (CVPR)}, 2023, pp. 739--748.

\bibitem{direct}
L.~Kneip and S.~Lynen, ``Direct optimization of frame-to-frame rotation,'' in \emph{Proc. IEEE Int. Conf. Comput. Vis. (ICCV)}, 2013, pp. 2352--2359.

\bibitem{opengv}
L.~Kneip and P.~Furgale, ``{OpenGV}: A unified and generalized approach to real-time calibrated geometric vision,'' in \emph{Proc. IEEE Int. Conf. Robot Autom. (ICRA)}.\hskip 1em plus 0.5em minus 0.4em\relax IEEE, 2014, pp. 1--8.

\bibitem{yolov5}
\BIBentryALTinterwordspacing
G.~Jocher, A.~Chaurasia, A.~Stoken, J.~Borovec, NanoCode012, Y.~Kwon, K.~Michael, TaoXie, J.~Fang, imyhxy, Lorna, Y.~Zeng, C.~Wong, A.~V, D.~Montes, Z.~Wang, C.~Fati, J.~Nadar, Laughing, and M.~Jain., ``{ultralytics/yolov5: v7.0 - {YOLOv5 SOTA} Realtime Instance Segmentation (v7.0)},'' Aug. 2022. [Online]. Available: \url{https://doi.org/10.5281/zenodo.7347926}
\BIBentrySTDinterwordspacing

\bibitem{deepsort}
N.~Wojke, A.~Bewley, and D.~Paulus, ``Simple online and realtime tracking with a deep association metric,'' in \emph{IEEE Int. Conf. Inf. Process. (ICIP)}.\hskip 1em plus 0.5em minus 0.4em\relax IEEE, 2017, pp. 3645--3649.

\bibitem{plvs}
L.~Freda, ``{PLVS}: A {SLAM} system with points, lines, volumetric mapping, and 3d incremental segmentation,'' \emph{arXiv preprint arXiv:2309.10896}, 2023.

\bibitem{euroc}
M.~Burri, J.~Nikolic, P.~Gohl, T.~Schneider, J.~Rehder, S.~Omari, M.~W. Achtelik, and R.~Siegwart, ``The {EuRoC} micro aerial vehicle datasets,'' \emph{Int J Rob Res}, vol.~35, no.~10, pp. 1157--1163, 2016.

\bibitem{asmle}
D.~Zu{\~n}iga-No{\"e}l, F.-A. Moreno, and J.~Gonzalez-Jimenez, ``An analytical solution to the {IMU} initialization problem for visual-inertial systems,'' \emph{IEEE Robot. Autom. Lett.}, vol.~6, no.~3, pp. 6116--6122, 2021.

\bibitem{dmvio}
L.~Von~Stumberg and D.~Cremers, ``{DM-VIO}: Delayed marginalization visual-inertial odometry,'' \emph{IEEE Robot. Autom. Lett.}, vol.~7, no.~2, pp. 1408--1415, 2022.

\bibitem{rdvio}
J.~Li, X.~Pan, G.~Huang, Z.~Zhang, N.~Wang, H.~Bao, and G.~Zhang, ``{RD-VIO}: Robust visual-inertial odometry for mobile augmented reality in dynamic environments,'' \emph{arXiv preprint arXiv:2310.15072}, 2023.

\bibitem{dynamslam}
H.~Yin, S.~Li, Y.~Tao, J.~Guo, and B.~Huang, ``{Dynam-SLAM}: An accurate, robust stereo visual-inertial {SLAM} method in dynamic environments,'' \emph{IEEE Trans. Robot.}, vol.~39, no.~1, pp. 289--308, 2022.

\bibitem{dvislam}
X.~Peng, Z.~Liu, W.~Li, P.~Tan, S.~Cho, and Q.~Wang, ``{DVI-SLAM}: A dual visual inertial {SLAM} network,'' \emph{arXiv preprint arXiv:2309.13814}, 2023.

\bibitem{mavis}
Y.~Wang, Y.~Ng, I.~Sa, A.~Parra, C.~Rodriguez, T.~J. Lin, and H.~Li, ``{MAVIS}: Multi-camera augmented visual-inertial {SLAM} using {SE2 (3)} based exact {IMU} pre-integration,'' \emph{arXiv preprint arXiv:2309.08142}, 2023.

\bibitem{openloris}
X.~Shi, D.~Li, P.~Zhao, Q.~Tian, Y.~Tian, Q.~Long, C.~Zhu, J.~Song, F.~Qiao, L.~Song \emph{et~al.}, ``Are we ready for service robots? the openloris-scene datasets for lifelong {SLAM},'' in \emph{Proc. IEEE Int. Conf. Robot Autom. (ICRA)}.\hskip 1em plus 0.5em minus 0.4em\relax IEEE, 2020, pp. 3139--3145.

\bibitem{tumvi}
D.~Schubert, T.~Goll, N.~Demmel, V.~Usenko, J.~St{\"u}ckler, and D.~Cremers, ``The {TUM VI} benchmark for evaluating visual-inertial odometry,'' in \emph{Proc. IEEE Int. Conf. Intel. Robots and Syst. (IROS)}.\hskip 1em plus 0.5em minus 0.4em\relax IEEE, 2018, pp. 1680--1687.

\end{thebibliography}



\end{document}